\documentclass[preprint,12pt]{elsarticle}

\usepackage{amssymb}
\usepackage{amsmath}
\usepackage{url}
\usepackage{comment}
\usepackage{microtype}
\usepackage{amsthm}
\theoremstyle{definition}
\newtheorem{definition}{Definition}
\journal{Information Fusion}

\usepackage{xcolor}

\newif\ifworkinprogress
\workinprogresstrue 
\ifworkinprogress

\else

\fi

\begin{document}

\begin{frontmatter}



\title{Orthogonalized Multimodal Contrastive Learning with Asymmetric Masking for Structured Representations}

\author[1]{Carolin Cissée}
\author[1,2]{Raneen Younis} 
\author[1,2]{Zahra Ahmadi\corref{cor1}} \ead{Ahmadi.Zahra@mh-hannover.de} 

\affiliation[1]{organization={Peter L. Reichertz Institute for Medical Informatics of TU Braunschweig and Hannover Medical School},
            city={Hannover},
            country={Germany}}
\cortext[cor1]{Corresponding authors}
\affiliation[2]{%
  organization={Lower Saxony Center for AI and 
Causal Methods in Medicine (CAIMed)},
  city={Hannover},
  country={Germany},
  }
\begin{abstract}
Multimodal learning seeks to integrate information from heterogeneous sources, where signals may be shared across modalities, specific to individual modalities, or emerge only through their interaction.
While self-supervised multimodal contrastive learning has achieved remarkable progress, most existing methods predominantly capture redundant cross-modal signals, often neglecting modality-specific (unique) and interaction-driven (synergistic) information. Recent extensions broaden this perspective, yet they either fail to explicitly model synergistic interactions or learn different information components in an entangled manner, leading to incomplete representations and potential information leakage.
We introduce \textbf{COrAL}, a principled framework that explicitly and simultaneously preserves redundant, unique, and synergistic information within multimodal representations. COrAL employs a dual-path architecture with orthogonality constraints to disentangle shared and modality-specific features, ensuring a clean separation of information components. To promote synergy modeling, we introduce asymmetric masking with complementary view-specific patterns, compelling the model to infer cross-modal dependencies rather than rely solely on redundant cues. 
Extensive experiments on synthetic benchmarks and diverse MultiBench datasets demonstrate that COrAL consistently matches or outperforms state-of-the-art methods while exhibiting low performance variance across runs. These results indicate that explicitly modeling the full spectrum of multimodal information yields more stable, reliable, and comprehensive embeddings\footnote{Code is available at: \url{https://github.com/Caro-Ci/COrAL}}.
\end{abstract}


\end{frontmatter}


\section{Introduction}
\label{sec:intro}
Multimodal learning aims to jointly model and integrate information from heterogeneous data sources, including images, text, audio, and structured data. By leveraging complementary cues across modalities, multimodal frameworks can produce more robust and context-aware representations than unimodal approaches. From an information-centric perspective, different modalities may encode (i) redundant information shared across sources, (ii) modality-specific signals unique to individual modalities, and (iii) synergetic information that emerges only through cross-modal interactions. Effectively capturing and integrating these distinct components is critical for reliable multimodal decision-making, with applications ranging from medical diagnosis and treatment support systems \cite{kodali2023multimodalAICancer, Zhang2025MultimodalAlzheimer, ElOuahidi2024MultimodalPacemaker} to autonomous driving \cite{hwang2024emmaMultimodal, Xiao2022AutoDriveMultimodal} and human–robot collaboration \cite{Wu2026robotmultimodal, duan2024multimodalrobot}.

Despite their promise, learning joint multimodal representations that faithfully preserve these heterogeneous information components remains challenging. Modalities differ substantially in structure, noise characteristics, and information density, and naïve fusion strategies often lead to information interference or loss. Approaches that overly prioritize shared representations may suppress modality-unique cues, whereas methods that treat modalities independently risk missing synergistic information arising from cross-modal interactions. These challenges are particularly pronounced in low-supervision settings, where models must primarily rely on intrinsic data structure rather than extensive labeled annotations \cite{Zhang2019ChallengesMultimodal, Deldari2022RepresentationLearning}.
Recent advances in self-supervised contrastive learning have substantially accelerated progress in multimodal representation learning \cite{Deldari2022RepresentationLearning, Zong2025Multimodal, Hu2024ContrastiveLearningOverviewUni, Jaiswal2021ContrastiveLearningUnimodal}. Building on unimodal frameworks such as SimCLR \cite{Chen20SimCLR}, MoCo \cite{He2020MoCo}, and DINO \cite{Caron2021DINO}, contrastive methods have been extended to multimodal settings, enabling heterogeneous inputs to be aligned within a shared embedding space \cite{Zong2025Multimodal, Hager2023ImgTabMulti, Radford2021CLIP}. This paradigm is particularly appealing in low-supervision regimes, as positive pairs can be constructed without manual labels through data augmentations or by modality pairing, with labels introduced only during downstream fine-tuning \cite{Deldari2022RepresentationLearning, Hu2024ContrastiveLearningOverviewUni, Jaiswal2021ContrastiveLearningUnimodal}. However, most existing contrastive multimodal frameworks primarily emphasize redundantly shared semantics across modalities, implicitly assuming that all task-relevant information resides within a single shared representation space \cite{Radford2021CLIP, Wang2025DECLIP, Zhu2021UniPerceiver, Tsai2021MultiViewRedundancy}. As a result, modality-specific features and synergic cross-modal interactions are often underrepresented or entangled, limiting robustness and generalization. Empirical evidence increasingly suggests that disentangling shared and unique information components is critical for improving downstream performance and stability.

To address these limitations, several recent approaches explicitly attempt to separate shared and modality-specific representations during training \cite{Liu2023FOCAL, Liang2023FactorCL, Wang2024DeCUR, Dufumier2024CoMM, wen2025infmasking}. 
Some methods further aim to model synergistic information arising from cross-modal interactions \cite{Dufumier2024CoMM, wen2025infmasking}. While these works demonstrate the importance of structured representation learning, information entanglement persists in practice, indicating that existing mechanisms may not enforce sufficiently strict separation between redundant, unique, and synergistic components.

In this work, we propose \textbf{COrAL}, a self-supervised \textbf{C}ontrastive \textbf{Or}thogonalized framework with \textbf{A}symmetric masking for \textbf{L}earning multimodal representations. COrAL explicitly models redundant, modality-unique, and synergistic information while enforcing a strict separation between shared and unique subspaces. The framework introduces a dual-path encoding strategy with orthogonality constraints to prevent information leakage between representation components, coupled with an asymmetric masking mechanism that promotes the emergence of synergistic information by partially obscuring individual modalities.
By design, COrAL overcomes key limitations of existing contrastive multimodal approaches, which often conflate shared and modality-specific signals or insufficiently capture cross-modal interactions \cite{Liu2023FOCAL, Liang2023FactorCL, Wang2024DeCUR, Dufumier2024CoMM, wen2025infmasking}. To the best of our knowledge, it is the first contrastive multimodal framework to jointly preserve redundant information, retain modality-specific cues, and actively promote synergistic representations under explicit orthogonal disentanglement constraints.

We evaluate COrAL on controlled synthetic data and multiple multimodal benchmarks. The results demonstrate improved recovery of modality-unique information relative to prior approaches \cite{Dufumier2024CoMM, wen2025infmasking}, while preserving redundant and synergistic components. This balanced representation learning translates into stable, competitive state-of-the-art performance with low variance across runs, suggesting robust optimization and improved generalization.

The remainder of this paper is organized as follows. Section \ref{sec:section_2} provides an overview of prior works. Section \ref{sec:section_3} formalizes the multimodal information decomposition problem. Section \ref{sec:section_4} presents the proposed COrAL framework and training objective. Section \ref{sec:section_5} reports empirical experiments on synthetic and MultiBench \cite{Liang2021MultiBench} datasets. Section \ref{sec:section_6} concludes with discussion and future directions. 
\section{Related work}
\label{sec:section_2}
\subsection{Self-supervised contrastive learning}
The scarcity of labeled data has driven the rapid development of self-supervised contrastive learning methods \cite{Hu2024ContrastiveLearningOverviewUni, Jaiswal2021ContrastiveLearningUnimodal}. These approaches learn representations without explicit supervision by constructing positive and negative pairs directly from the data. Typically, different augmentations of the same sample are treated as positive pairs, under the assumption that appropriately augmented views should preserve semantic consistency, while views of different samples form negative pairs whose representations are encouraged to diverge. By maximizing similarity between positive pairs and minimizing similarity between negative pairs, contrastive objectives enable the learning of semantically meaningful representations without manual annotation.
A key advantage of self-supervised contrastive learning lies in its label efficiency: the learned representations can be fine-tuned using limited labeled data for downstream tasks, often achieving competitive or superior performance compared to fully supervised approaches \cite{Hu2024ContrastiveLearningOverviewUni, Jaiswal2021ContrastiveLearningUnimodal, Dufumier2024CoMM}. This effectiveness has been demonstrated extensively in the visual domain. For instance, SimCLR \cite{Chen20SimCLR} showed that a simple contrastive framework with strong data augmentations can achieve performance comparable to supervised learning on unimodal data. MoCo \cite{He2020MoCo} further improved scalability by introducing a momentum-based dynamic dictionary, enabling contrastive learning with large and consistent sets of negative samples. Alternative formulations have reduced the reliance on explicit negative pairs through mechanisms such as momentum encoders or gradient stopping, as exemplified by methods such as DINO \cite{Caron2021DINO}, BYOL \cite{Grill2020BYOL}, and SimSiam \cite{Chen2020SimSiam}.
While these approaches have firmly established self-supervised contrastive learning as a powerful paradigm for representation learning, they primarily focus on unimodal settings. Extending these principles to multimodal data introduces additional challenges related to modality heterogeneity and information interaction, which are not addressed by unimodal contrastive frameworks.

\subsection{Multimodality in self-supervised contrastive learning}
The success of self-supervised contrastive learning in unimodal settings has motivated its extension to multimodal data, aiming to jointly model heterogeneous modalities and their interactions. In multimodal contrastive learning, positive pairs are extended to include cross-modal correspondences, enabling alignment across modalities in a shared representation space. A representative example is CLIP \cite{Radford2021CLIP}, which learns aligned image–text representations by treating images and their corresponding captions as positive pairs.
Subsequent work has focused on improving the efficiency and effectiveness of such frameworks. DECLIP \cite{Wang2025DECLIP}, for instance, enriches positive pair construction through augmented intra-modal views, multiple cross-modal views per sample, and nearest-neighbor positives. Despite these advances, many multimodal contrastive approaches remain tailored to specific modality combinations, limiting their general applicability \cite{Liu2023FOCAL, Akbari2021VATT, Wang2023COMET, Liu2024TimesURL}. To address this, modality-agnostic architectures such as Uni-Perceiver \cite{Zhu2021UniPerceiver} employ a unified Transformer encoder with modality-specific tokenizers to support diverse modalities and tasks.

\subsection{Redundancy, synergy and uniqueness}
Most multimodal self-supervised contrastive learning approaches focus solely on extracting information that is shared across modalities, implicitly assuming that task-relevant information is redundantly encoded in each modality and can therefore be recovered through cross-modal alignment \cite{Akbari2021VATT, Tsai2021MultiViewRedundancy, Tosh2021ContrastiveMultiview, Radford2021CLIP, Tian2020ViewsRedundancy}. While effective for learning shared semantics, this assumption neglects modality-specific signals and limits the ability of such methods to capture richer forms of multimodal information.

Recent work has therefore sought to extend multimodal representations beyond shared information by explicitly modeling modality-unique components \cite{Liu2023FOCAL, Liang2023FactorCL, Wang2024DeCUR, Dufumier2024CoMM}. These approaches distinguish between shared and unique information and introduce architectural or objective-level mechanisms to preserve both. FOCAL \cite{Liu2023FOCAL}, for instance, constructs orthogonal latent subspaces for shared and modality-specific features and applies distinct contrastive objectives within each subspace. DeCUR \cite{Wang2024DeCUR} similarly separates shared and unique information, but does so implicitly by partitioning a single latent space without enforcing explicit orthogonality. FactorCL \cite{Liang2023FactorCL} adopts an information-theoretic perspective, factorizing task-relevant information into shared and unique components by jointly maximizing lower bounds on task-relevant information and minimizing upper bounds on task-irrelevant information using contrastive estimators.

Beyond shared and unique information, CoMM \cite{Dufumier2024CoMM} further models synergistic information that emerges through cross-modal interactions. It aligns multimodal representations using complementary contrastive objectives that capture redundant, unique, and synergistic components, and employs exchangeable modality-specific encoders to support diverse modality combinations. Wen et al. \cite{wen2025infmasking} extend this framework with InfMasking, which enhances synergy capture by stochastically masking modality features and pairing partial and complete representations, approximating infinitely many masked views.
Despite these advances, existing approaches do not simultaneously enforce strict disentanglement between shared and modality-specific information while actively promoting synergistic representations. In particular, CoMM and InfMasking do not explicitly separate shared and unique components in the representation space, which may lead to information interference and reduced interpretability in downstream tasks.

In contrast, COrAL learns multimodal representations that explicitly capture redundant, modality-unique, and synergistic information within disentangled shared and unique subspaces. Unlike prior methods, COrAL enforces orthogonality to strictly separate shared and modality-specific components and captures them in isolation through a dual-path architecture. Moreover, while both COrAL and InfMasking use partial masking to enhance synergy capture, COrAL integrates asymmetric masking directly into the shared pathway, eliminating the need for additional representations.

\section{Problem formulation}
\label{sec:section_3}

Let $X = (X_1, \ldots, X_n)$ denote a multimodal input comprising $n \geq 2$ modalities, and let $Y$ denote the target variable. Following the Partial Information Decomposition (PID) framework\cite{Williams2010PID}, the mutual information between $X$ and $Y$ can be decomposed into conceptually distinct components that reflect different modes of information contributions.
To make this decomposition explicit, we first consider the bimodal case $X = (X_1, X_2)$, which provides an interpretable view of the underlying information components. In this setting, the mutual information admits the decomposition
\begin{equation}
  I(X; Y) = R + S + U_1 + U_2,
  \label{eq:mutualinformation}
\end{equation}
where $R$, $S$, and $U_i$ represent redundant, synergistic, and modality-unique information, respectively. We define these components as follows:
\begin{definition}[Redundancy]
\label{def:redundancy}
The \emph{redundancy} $R$ is the information about $Y$ that is shared by both modalities $X_1$ and $X_2$.
\end{definition}

\begin{definition}[Synergy]
\label{def:synergy}
The \emph{synergy} $S$ is the information about $Y$ that arises only through the joint cross-modal interactions of $X_1$ and $X_2$, and cannot be attributed to either modality in isolation. 
\end{definition}

\begin{definition}[Uniqueness]
\label{def:unique}
For modality $i \in \{1, 2\}$, the \emph{uniqueness} $U_i$ is the information about $Y$ contained exclusively in modality $X_i$.
\end{definition}

While the bimodal case offers a clear view, PID naturally extends to general multimodal inputs with $n \geq 2$ modalities. In the general setting, redundancy and synergistic information are organized over a lattice of partial information atoms \cite{Williams2010PID}, capturing interactions ranging from pairwise modality combinations to higher-order dependencies involving multiple modalities. In this work, we adopt a simplified principled notation in which $R$ and $S$ denote aggregate redundancy and synergy across all interaction orders. This abstraction preserves the interpretability central to PID while avoiding the combinatorial complexity of the full lattice representation.

From this perspective, an ideal multimodal representation should preserve all information components—redundant, synergistic, and modality-unique—to retain the full spectrum of task-relevant information. To this end, COrAL constructs embeddings in disentangled representation subspaces via a dual-path design:
\begin{itemize}
    \item $Z^{SR} = F^{SR}(X)$, which captures shared and synergistic information ($R + S$),
    \item $Z^U_i = F^U_i(X_i)$, which preserves modality-unique information $U_i$ for $i \in \{1,\ldots,n\}$.
\end{itemize}
The final representation is formed by concatenation, 
\begin{equation}
  Z = (Z^{SR}, Z^U_1, \ldots, Z^U_n),
  \label{eq:concatembedding}
\end{equation}
and is designed to jointly encode the complementary information components present in multimodal observations.

\section{COrAL framework}
\label{sec:section_4}

\begin{figure*}[t]
    \centering
\includegraphics[width=1\textwidth]{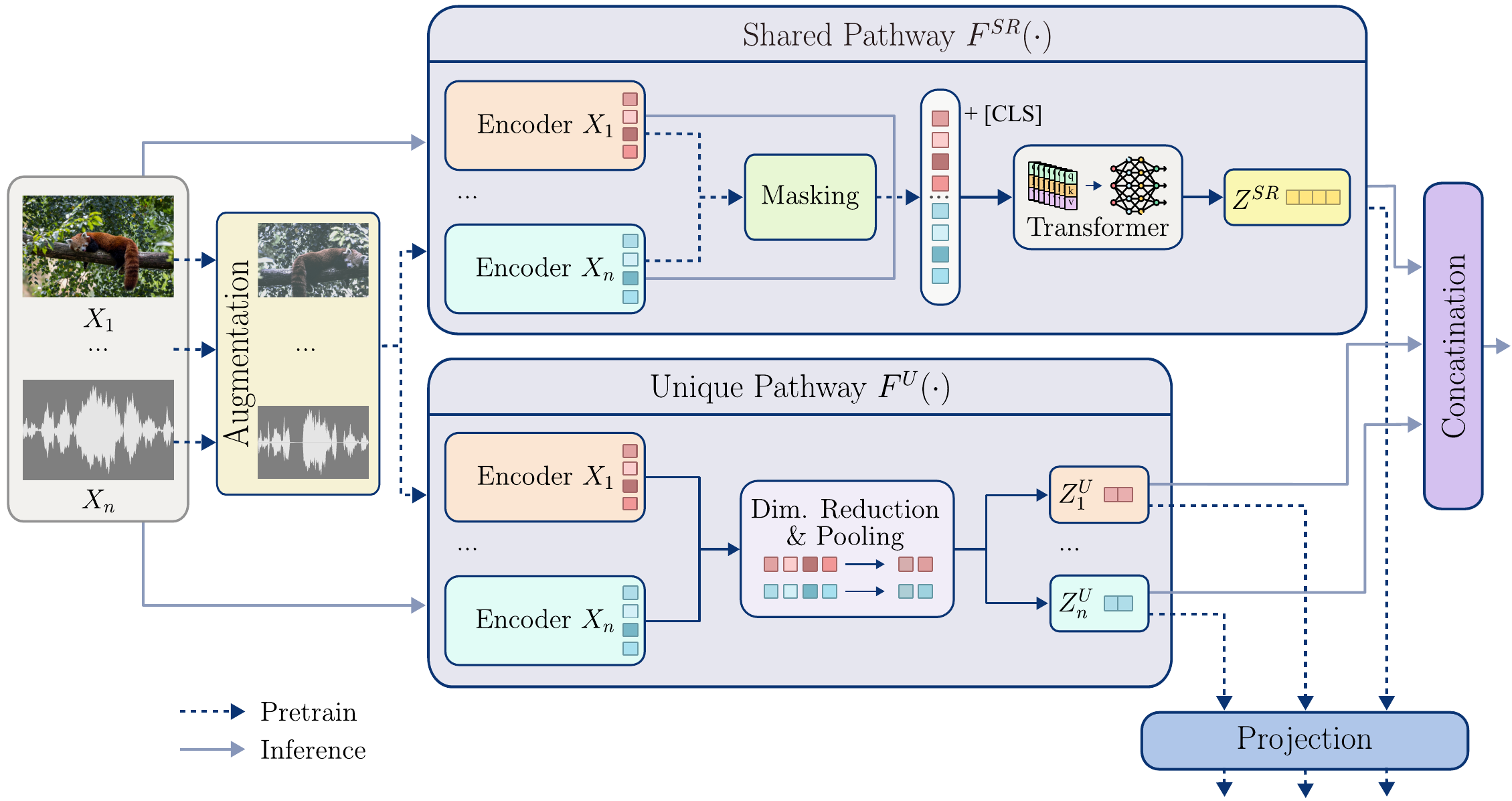}
    \caption{COrAL Architecture for multimodal input $X=(X_1,\ldots, X_n)$. Each modality is processed by a shared pathway $F^{SR}(\cdot)$ for an embedding $Z^{SR}$, which captures synergistic and redundant information, and by a modality-specific unique pathway $F^{U}(\cdot)$ to produce $Z^{U}_1,\ldots,Z^{U}_n$. At inference, neither augmentation nor masking is applied.}
    \label{fig:figure-1}
\end{figure*}

COrAL is designed to learn comprehensive multimodal representations by explicitly modeling redundant, modality-unique, and synergistic information. Motivated by evidence that different modalities contribute complementary cues to a common task \cite{Liu2023FOCAL, Liang2023FactorCL, Wang2024DeCUR, Dufumier2024CoMM}, COrAL combines a dual-path architecture with an asymmetric masking strategy to disentangle and align multimodal information in a principled manner.

\subsection{Dual-path architecture with asymmetric masking}
Figure~\ref{fig:figure-1} provides an overview of the COrAL framework, a transformer-based multimodal architecture explicitly structured to capture redundancy ($R$), synergy ($S$), and modality-unique components ($U_i$). Building on multimodal transformer designs \cite{Liu2023FOCAL, Dufumier2024CoMM}, COrAL integrates modality-specific encoders with a fusion transformer and introduces two tightly coupled mechanisms: (i) a dual-path encoder architecture and (ii) an asymmetric masking strategy.

For each input modality, COrAL instantiates two parallel encoding pathways. The shared pathway extracts a multimodal representation $Z^{SR}$ that captures redundant and synergistic information across modalities. In parallel, the unique pathway encodes modality-specific representations $Z^U_i$ that preserve information unique to each modality. Each pathway employs modality-specific encoders; when required, a latent converter maps encoder outputs into token sequences compatible with the fusion transformer.
During pretraining, asymmetric masking is applied within the shared pathway after the modality-specific encoding (and latent conversion, when used) and before fusion. By selectively masking modality features in complementary views, the model is encouraged to maintain embedding consistency under structured information gaps, thereby promoting inference from cross-modal dependencies rather than relying solely on redundant signals. The masked token sequences from all modalities are concatenated and processed by the fusion transformer, yielding the shared representation.

In contrast, each unique pathway processes its corresponding modality independently. After encoding and tokenization, the representation dimensionality is reduced to half of the shared embedding size, and attention pooling produces a compact and discriminative modality-specific vector representation. A separate unique pathway is maintained for each input modality.
For pretraining, all representations are projected into a common embedding space through projection heads, and contrastive objectives are applied to the projected embeddings. This design enables joint optimization of shared and modality-unique representations within a unified training framework.

\subsection{Training objective} 
During training, two augmentations $t^\prime, t^{\prime\prime} \sim T$ are applied to the multimodal input $X = (X_1, \ldots, X_n)$, producing two correlated views $X^\prime$ and $X^{\prime\prime}$. Each view is processed by the dual-path network $F(\cdot)$, comprising a shared pathway $F^{SR}(\cdot)$ and modality-specific unique pathways $F^U(\cdot)$, yielding embeddings $\{Z^{SR\prime}, Z^{U\prime}_1, \ldots, Z^{U\prime}_n\}$ and $\{Z^{SR\prime\prime}, Z^{U\prime\prime}_1, \ldots, Z^{U\prime\prime}_n\}$, respectively.
The overall COrAL objective comprises three terms:
\begin{equation}
    \mathcal{L}_{\text{COrAL}} =\lambda_s \cdot \mathcal{L}_{\text{shared}} + \lambda_u \cdot\mathcal{L}_{\text{unique}} + \lambda_o \cdot\mathcal{L}_{\text{orthogonal}},
  \label{eq:overallloss}
\end{equation}
where $\lambda_s$, $\lambda_u$, and $\lambda_o$ control the relative contribution of each term.
The shared loss $\mathcal{L}_{\text{shared}}$ enforces contrastive alignment in the shared pathway $F^{SR}(\cdot)$, capturing both redundant and synergistic information across modalities. The unique loss $\mathcal{L}_{\text{unique}}$ applies contrastive learning independently within each modality-specific pathway $F^U(\cdot)$ to preserve modality-unique information. Both objectives are implemented using the InfoNCE estimator \cite{Oord2019InfoNCE}, widely used in contrastive representation learning \cite{He2020MoCo, Radford2021CLIP, Liu2023FOCAL, Liang2023FactorCL, Dufumier2024CoMM, wen2025infmasking}:

\begin{align}
\begin{split}
    \mathcal{L}_{\text{shared}} &= - \hat{I}_{\text{NCE}}(Z^{SR\prime},Z^{SR\prime\prime}), \\
    \mathcal{L}_{\text{unique}} &= - \sum^n_{i=1} \hat{I}_{\text{NCE}} (Z^{U\prime}_{i} ,Z^{U\prime\prime}_{i} ), \\
      \hat{I}_{\text{NCE}}(Z, Z^\prime) &= \mathbb{E}_{\substack{z,z^\prime_{\text{pos}} \sim \hat{p}(Z,Z^\prime) \\ z^\prime_{\text{neg}} \sim \hat{p}(Z^\prime)}} \left[ \log \frac{\exp\text{sim}(z, z^\prime_{\text{pos}})}{\sum_{z^\prime_{\text{neg}}} \exp\text{sim}(z, z^\prime_{\text{neg}})} \right]. 
  \label{eq:infonce}
\end{split}
\end{align}

To explicitly promote the learning of synergistic information, COrAL incorporates an asymmetric masking operation $M_m(\cdot)$ within the shared pathway $F^{SR}(\cdot)$. The masking ratio $m$ is gradually increased during training. Masking is applied after modality-specific encoding and latent conversion, and before fusion by the transformer $T(\cdot)$. For bimodal inputs, the shared embeddings are defined as:
\begin{eqnarray}
     Z^{SR\prime} = T(M_m(E^{SR}_1(X^\prime_1)),E^{SR}_2(X^\prime_2)),\nonumber\\
     Z^{SR\prime\prime} = T(E^{SR}_1(X^{\prime\prime}_1),M_m(E^{SR}_2(X^{\prime\prime}_2))).
\end{eqnarray}

\begin{figure}
    \centering
    \includegraphics[width=0.45\linewidth]{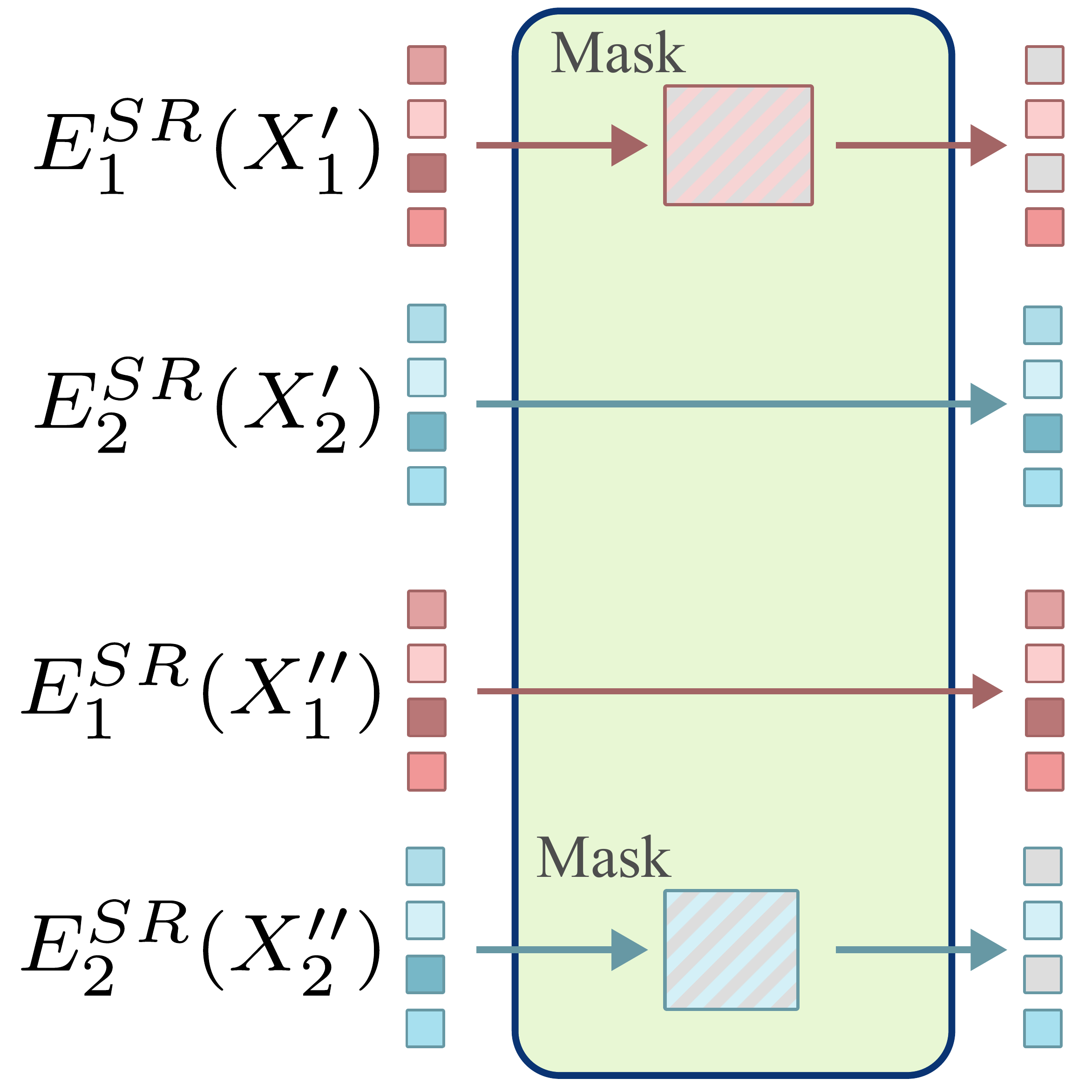}
    \caption{COrAL asymmetric masking strategy for two views $X^\prime$ and $X^{\prime\prime}$ of a bimodal input $X = (X_1, X_2)$ after encoding and latent conversion $E^{SR}_i(\cdot)$.}
    \label{fig:figure-2}
\end{figure}

Figure~\ref{fig:figure-2} illustrates this mechanism for a bimodal case: one view masks features from the first modality, while the complementary view masks the second modality. 
For general multimodal inputs with $n \geq 2$ modalities, this strategy is extended by randomly partitioning modalities into two disjoint subsets $\mathcal{M}_1$ and $\mathcal{M}_2$ of equal size for each batch. Masking is applied to $\mathcal{M}_1$ in the first view and to $\mathcal{M}_2$ in the second view:
\begin{eqnarray}
     Z^{SR\prime} = T\left(\{M_m(E^{SR}_i(X^\prime_i))\}_{i \in \mathcal{M}_1}, \{E^{SR}_j(X^\prime_j)\}_{j \in \mathcal{M}_2}\right),\nonumber\\
     Z^{SR\prime\prime} = T\left(\{E^{SR}_i(X^{\prime\prime}_i)\}_{i \in \mathcal{M}_1}, \{M_m(E^{SR}_j(X^{\prime\prime}_j))\}_{j \in \mathcal{M}_2}\right).
\end{eqnarray}

This structured occlusion prevents the model from trivially matching view through redundant signals alone. Instead, maintaining shared embedding consistency under complementary masking encourages the model to leverage cross-modal dependencies, providing a direct learning signal for synergistic information that might otherwise be dominated by redundancy.

The orthogonality loss $\mathcal{L}_{\text{orthogonal}}$ enforces a strict separation between the shared embedding $Z^{SR}$ and the modality-unique embeddings $Z^U_i$, ensuring that each subspace captures independent semantic information and preventing information leakage across representation components. To this end, we adopt the cosine embedding loss (CEL), also used in FOCAL \cite{Liu2023FOCAL}, to encourage orthogonality between embeddings:
\begin{equation}
     \text{CEL}(Z, Z^\prime) = \mathbb{E}_{z,z^\prime \sim \hat{p}(Z,Z^\prime)} \left[ \max\left(0, \text{cos\_sim}\left(z, z^\prime\right)\right) \right]. 
\label{eq:cosineloss}
\end{equation}

The cosine embedding loss is applied to enforce separation between the shared embedding $Z^{SR}$ and the modality-unique embeddings $\{Z^U_1, \ldots, Z^U_n\}$ within each view, as well as between different modality-unique embeddings. Specifically, the orthogonality loss is defined as:
\begin{align}
\mathcal{L}_{\text{orthogonal}} = &\sum_{i=1}^n\underbrace{\left(\text{CEL}(Z^{U\prime}_i, Z^{SR\prime}) + \text{CEL}(Z^{U\prime\prime}_i, Z^{SR\prime\prime})\right)}_{\text{U\(_i\) vs SR}} + \nonumber\\
&\sum_{i=1}^{n-1} \sum_{j=i+1}^n\underbrace{\left(\text{CEL}(Z^{U\prime}_i, Z^{U\prime}_j) + \text{CEL}(Z^{U\prime\prime}_i, Z^{U\prime\prime}_j)\right)}_{\text{U\(_i\) vs U\(_j\)}}. 
\label{eq:ortho_loss}
\end{align}

\begin{figure}[t]
    \centering
    \includegraphics[width=0.5\textwidth]{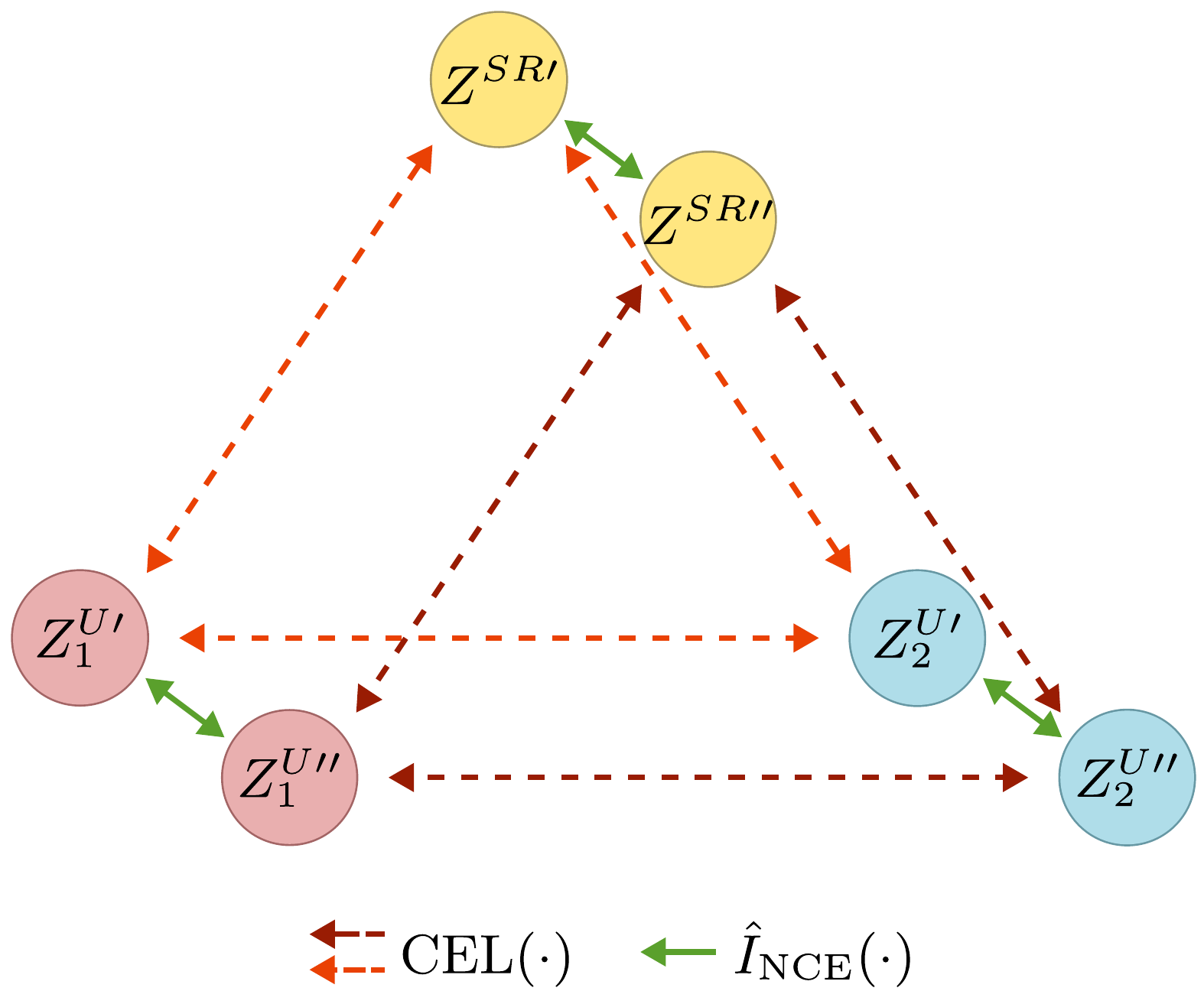}
    \caption{Schematic of the loss calculations between the embeddings of two views $X^\prime$ and $X^{\prime\prime}$ for bimodal input, as defined by the loss components of COrAL. $Z^{SR\prime}$ and $Z^{SR\prime\prime}$ are the embeddings containing information shared between modalities and $Z^{U\prime}_1$, $Z^{U\prime}_2$, $Z^{U\prime\prime}_1$, and $Z^{U\prime\prime}_2$ are the embeddings of modality-unique information. Arrows with $\text{CEL}(\cdot)$ indicate that cosine embedding loss is calculated between embeddings, and arrows with $\hat{I}_{\text{NCE}}(\cdot)$ refer to the InfoNCE estimator of mutual information.}
    \label{fig:figure-3}
\end{figure}

Unlike squared cosine similarity losses that enforce exact orthogonality, $\mathcal{L}_{\text{orthogonal}}$ allows angles of at least $90^\circ$ between embeddings. This relaxation improves optimization flexibility while preserving effective separation of semantic information across subspaces.
Figure~\ref{fig:figure-3} summarizes the loss structure for the bimodal case. The shared loss $\mathcal{L}_{\text{shared}}$ aligns $Z^{SR\prime}$ and $Z^{SR\prime\prime}$, the unique loss $\mathcal{L}_{\text{unique}}$ aligns each modality-specific embedding pair $(Z^{U\prime}_i, Z^{U\prime\prime}_i)$, and the orthogonality loss enforces within-view separation between shared and unique representations.

During inference, neither data augmentation nor masking is applied. The multimodal input $X = (X_1, \ldots, X_n)$ is processed through the shared and unique pathways, $F^{SR}(\cdot)$ and $F^U(\cdot)$, and the resulting embeddings $(Z^{SR}, Z^U_1, \ldots, Z^U_n)$ are concatenated to form the final representation $Z$.

\section{Experiments}
\label{sec:section_5}
We evaluate the proposed COrAL framework on both controlled synthetic data and diverse multimodal benchmark datasets. Our experimental design serves two objectives: (i) to assess whether COrAL learns disentangled representations that preserve redundant, modality-unique, and synergistic information, and (ii) to examine how these properties translate into performance on real-world datasets.

\subsection{Comparison methods}

We compare COrAL against four representative state-of-the-art multimodal contrastive learning approaches. CoMM \cite{Dufumier2024CoMM} employs a multimodal transformer to model redundant, unique, and synergistic information. InfMasking \cite{wen2025infmasking} extends CoMM by introducing a masking strategy to enhance the capture of synergistic information. FactorCL \cite{Liang2023FactorCL} applies factorized contrastive learning to disentangle shared and modality-specific information. Finally, CLIP \cite{Radford2021CLIP} serves as a standard cross-modal contrastive learning baseline that primarily captures redundantly shared semantics without modeling unique or synergistic information.
While FactorCL, CoMM, and InfMasking are explicitly designed to decompose multiple information components, CLIP provides a competitive baseline for redundancy-focused representations.

\subsection{Datasets}
We evaluate COrAL’s ability to capture redundant, synergistic, and modality-unique information using the synthetic bimodal Trifeature dataset introduced by Dufumier \textit{et al.} \cite{Dufumier2024CoMM}. This controlled synthetic setting enforces the presence of each information component through carefully designed data generation and pairing strategies. 
The dataset consists of pairs of generated images forming two modalities. Each image depicts a geometric shape characterized by three independent attributes: shape, texture, and color, each taking one of ten discrete values. Information components are induced through specific pairing strategies: Redundancy is enforced by pairing images that share the same shape across modalities, while texture remains a modality-unique feature. Synergy is defined via an artificial mapping between the texture of the first modality and the color of the second modality, with training pairs constructed to follow this mapping. 
The dataset contains 10,000 training pairs and 4,096 test pairs, where test pairs preserve shape consistency while varying texture. Synergy capture is evaluated by linear probing to predict adherence to the predefined texture–color mapping. Redundancy is assessed by predicting the shared shape across modalities, and uniqueness by predicting the texture of the first modality. Random guessing yields an accuracy of 10\% for redundancy and uniqueness tasks and 50\% for the synergy task.
We further evaluate COrAL on five benchmark datasets from MultiBench \cite{Liang2021MultiBench}:
\begin{itemize}
    \item \textbf{MIMIC III} \cite{Johnson2016MIMIC3} contains medical records data of over 40,000 ICU patients in the form of tabular patient data and time-series of ICU monitoring parameters. The data that is provided by MultiBench is preprocessed and contains 5 numerical values with patient data and 12 numerical measurements taken every hour over a 24-hour period. The classification goal is whether a patient has a respiratory disease during their stay, determined by any ICD-9 code in group 7 (ICD-9 codes 460-519).
    \item \textbf{CMU-MOSEI} \cite{Zadeh2018MOSEI}, a multimodal sentiment and emotion benchmark dataset of 23,000 monologue videos from over 1,000 speakers and 250 topics. Overall, the videos exceed 65 hours in length. We use the extracted features of the vision and text modalities from MultiBench. The sentiment intensity is labeled with numerical values between -3 and 3; however, prior work \cite{Liang2023FactorCL, Dufumier2024CoMM} uses a binary classification of positive or negative sentiment instead.
    \item \textbf{CMU-MOSI} \cite{Zadeh2016MOSI} is a dataset for multimodal sentiment analysis from 2,199 YouTube clips. Similar to CMU-MOSEI, sentiment intensity is labeled continuously from -3 to 3. Again, following prior work \cite{Liang2023FactorCL, Dufumier2024CoMM}, we consider the binary classification (sentiment negative/positive), and we use the extracted vision and text features provided by MultiBench.
    \item \textbf{UR-FUNNY} \cite{Hasan2019URFUNNY}, a dataset for humor detection in speech, consists of samples from more than 16,000 TED talk videos. It contains 8,257 humorous punchlines, including the sentences that precede them, and the same number of negative samples without humorous remarks. The task is to classify whether humour is present in a clip, based on MultiBench's visual and text features.
    \item \textbf{MUsTARD} \cite{Castro2019MUsTARD}, a dataset for sarcasm detection with 690 videos from popular television shows, including the previous sentences for context and an utterance for the binary classification task of sarcasm with text and visual features. 
\end{itemize}

\subsection{Implementation details}
To ensure fair comparison, all models are trained using identical encoder backbones for each dataset and are evaluated under the same experimental protocol. Each experiment is repeated five times with different random seeds (41–45), and we report the mean accuracy and standard deviation across runs. All models are trained for 100 epochs and evaluated via linear probing on the pretrained representations.

For COrAL, we use the AdamW optimizer with a learning rate of $1\times10^{-4}$, weight decay of $1\times10^{-3}$, and a batch size of 64. Unless stated otherwise, the loss coefficients $\lambda_s$, $\lambda_u$, and $\lambda_o$ are set to 1. To progressively encourage the learning of synergistic information, the asymmetric masking ratio increases from 5\% to 75\% in four stages over 100 training epochs.
Baseline methods are trained using a fixed set of hyperparameters reported in their original implementations. Specifically, we use the following hyperparameters: $1\times 10^{-4}$ learning rate for FactorCL; $1\times 10^{-3} $ learning rate, $1 \times 10^{-2}$ weight decay, and $0.1$ temperature for CoMM; $3 \times 10^{-4}$ learning rate, $0.1$ temperature, $0.7$ masking ratio, and 6 masked views for InfMasking; and $1 \times 10^{-3}$ learning rate and $1 \times 10^{-2}$ weight decay for CLIP. This allows us to isolate the effects of architectural and objective-level differences, rather than hyperparameter optimization, across all experiments.

For the Trifeature dataset, we follow \cite{Dufumier2024CoMM} and use AlexNet without the final average pooling layer as the visual encoder backbone. 
Feature maps are converted into token sequences via a linear patch embedding layer. For benchmark datasets, modality-specific encoders follow standard configurations: Transformers for text and vision, a shallow MLP with a feature tokenizer for tabular data, and a gated recurring unit for time-series data. For all experiments, a lightweight Transformer with eight attention heads and an additional [CLS] learnable embedding is used as a fusion module in the shared pathway to integrate modality-specific representations, while modality-specific unique pathways use a 2-layer MLP with attention pooling.

\subsection{Results on synthetic data}
Table~\ref{tab:table-1} summarizes results on the synthetic Trifeature dataset. All methods capture redundancy effectively (99-100\% accuracy). As expected, CLIP achieves perfect redundancy performance but fails to capture modality-unique and synergistic information, reflecting the limitations of standard cross-modal contrastive objectives.
FactorCL improves uniqueness modeling but remains near chance performance on synergy. In contrast, CoMM, InfMasking, and COrAL achieve above-chance performance across all components. InfMasking attains the highest synergy accuracy (77.0\%), while COrAL achieves a closely comparable 74.4\%, indicating that asymmetric masking strategy effectively enhances synergy capture beyond CoMM.
In addition, COrAL exhibits a clear improvement in capturing modality-unique information (96.4\%) while maintaining low variance across runs. This results highlight the effectiveness of its orthogonalized representation space and dual-path encoder in disentangling modality-specific and shared factors.
 \begin{table}[t]
\centering
\begin{tabular}{lccc}
\hline
\textbf{Model}&\textbf{Redundancy}  &\textbf{Uniqueness}  & \textbf{Synergy} \\
\hline
CLIP \cite{Radford2021CLIP}* & $\mathbf{100.0}$ & $11.6$ & $50.0$\\
FactorCL \cite{Liang2023FactorCL}* & $99.8$ & $62.5$ & $46.5$\\
CoMM \cite{Dufumier2024CoMM}* & $99.9$ & $87.8$ & $71.9$ \\
InfMasking \cite{wen2025infmasking}**& ${99.9}_{\pm0.09}$ & $90.6_{\pm2.31}$ & $\mathbf{77.0}_{\pm4.22}$ \\
COrAL (ours)& $99.7_{\pm0.08}$ & $\mathbf{96.4}_{\pm1.08}$ & ${74.4_{\pm2.92}}$\\
\hline
\end{tabular}
\caption{Linear probing accuracy (in \%) on the bimodal Trifeature dataset. The best results are marked in bold. * denotes results are taken from \cite{Dufumier2024CoMM}. ** denotes results are taken from \cite{wen2025infmasking}.}
\label{tab:table-1}

\end{table}
\subsection{Results on MultiBench datasets}

\begin{table*}[t]
\centering
\resizebox{\textwidth}{!}{
\begin{tabular}{lcccccc}
\hline
\textbf{Model} & \textbf{MIMIC} & \textbf{MOSEI} & \textbf{MOSI} & \textbf{UR-FUNNY} & \textbf{MUsTARD} & \textbf{Average} \\
\hline
CLIP\cite{Radford2021CLIP} & $66.0_{\pm0.39}$ & $64.2_{\pm0.45}$ & $50.6_{\pm0.75}$& $57.4_{\pm1.25}$ & $65.5_{\pm2.71}$ & $60.7_{\pm5.94}$\\
FactorCL\cite{Liang2023FactorCL} & $65.1_{\pm0.21}$ & $\mathbf{71.7}_{\pm1.38}$ & $52.5_{\pm1.74}$& $59.8_{\pm0.77}$ & $54.2_{\pm2.40}$ & $60.7_{\pm7.08}$\\
CoMM \cite{Dufumier2024CoMM} & $65.7_{\pm0.42}$ & $69.2_{\pm0.48}$ & $66.0_{\pm1.67}$ & $\underline{63.0}_{\pm1.08}$ & $\underline{65.3}_{\pm2.21}$ & $65.8_{\pm1.98}$ \\
InfMasking \cite{wen2025infmasking}& $\mathbf{67.7}_{\pm0.37}$ & $69.7_{\pm0.23}$ & $\underline{67.2}_{\pm1.00}$& $62.8_{\pm1.22}$ & $\mathbf{65.5}_{\pm0.62}$ & $\underline{66.6}_{\pm2.32}$\\
COrAL (ours) & $\underline{67.4}_{\pm0.22}$ & $\underline{70.4}_{\pm0.21}$ & $\mathbf{67.9}_{\pm0.70}$ & $\mathbf{63.5}_{\pm1.04}$& $64.4_{\pm1.76}$& ${\mathbf{66.7}_{\pm2.79}}$ \\
\hline
\end{tabular}
}
\caption{Linear evaluation accuracy (in \%) after 100 epochs of self-supervised learning on MultiBench datasets. The best results are marked in bold, the second-best results are underlined.}\label{tab:table-2}
\end{table*}

Table~\ref{tab:table-2} reports linear probing results on the MultiBench datasets after self-supervised pretraining. Overall, COrAL demonstrates consistently strong performance across benchmarks. It outperforms CoMM on all five datasets, CLIP on four (MIMIC, MOSEI, MOSI, and UR-FUNNY), and FactorCL on four (MIMIC, MOSI, UR-FUNNY, and MUsTARD), while FactorCL achieves the best performance on MOSEI. InfMasking attains the highest accuracy on MIMIC and matches CLIP on MUsTARD, while exhibiting lower variance across runs. Although the performance difference between COrAL and InfMasking on MIMIC is marginal (0.3\%).
In terms of average accuracy, COrAL achieves the highest overall performance, albeit by a narrow margin of 0.1\%. Importantly, COrAL consistently exhibits low variance across runs, achieving the lowest standard deviation on MOSEI, MOSI, and UR-FUNNY, indicating robust and reliably optimized learned representations across different random initializations.

Notably, CLIP performs competitively on MIMIC and MUsTARD despite modeling only redundant information, and FactorCL performs strongly on MOSEI, where redundancy and modality-specific cues appear sufficient for the task. This suggests that the relative importance of redundant, unique, and synergistic information varies across tasks, and that improvements in modeling unique and synergistic components do not necessarily translate into gains when redundancy alone is sufficient.
Overall, COrAL demonstrates stable and competitive performance across diverse multimodal benchmarks, outperforming prior methods on two datasets and achieving comparable results on the remaining benchmarks.

\subsection{Ablation studies}
\label{sec:ablation}
To rigorously validate the design of COrAL, we conduct a series of targeted ablation studies examining (i) the impact of asymmetric masking in synergy capture, (ii) sensitivity to the loss weights $\lambda_s$, $\lambda_u$, and $\lambda_o$, (iii) architectural design choices for the unique-information pathway, and (iv) the effectiveness of embedding space disentanglement. Together, these analyses provide both quantitative and qualitative evidence supporting the proposed design.

\subsubsection{Asymmetric masking}
\begin{table}[t]
\centering
\begin{tabular}{lccc}
\hline
\textbf{Masking}&\textbf{Redundancy}  &\textbf{Uniqueness}  & \textbf{Synergy} \\
\hline
Increasing ratio& $\mathbf{99.7}_{\pm0.08}$ & ${96.4_{\pm1.08}}$ & ${74.4_{\pm2.92}}$ \\
None & $99.6_{\pm0.22}$ & $93.9_{\pm2.70}$ & $50.7_{\pm0.38}$ \\
25\% masking & $99.7_{\pm0.14}$ & $96.6_{\pm1.39}$ & $74.8_{\pm1.25}$ \\
50\% masking & $99.5_{\pm0.08}$ & $\mathbf{97.4}_{\pm0.63}$ & $75.5_{\pm1.09}$ \\
75\% masking& $98.6_{\pm0.74}$ & $96.3_{\pm0.64}$ & $\mathbf{76.9}_{\pm1.99}$ \\
\hline
\end{tabular}
\caption{Linear probing accuracy (in \%) of COrAL on Trifeature after 100 epochs of self-supervised learning, comparing the effects of asymmetric masking with increasing amounts, static masking percentages, and no masking. Best results are marked in bold.}
\label{tab:table-3}
\end{table}

\begin{table*}[t]
\centering
\resizebox{\textwidth}{!}{
\begin{tabular}{lccccc}
\hline
\textbf{Masking} &\textbf{MIMIC}  &\textbf{MOSEI}  & \textbf{MOSI} &\textbf{UR-FUNNY} & \textbf{MUsTARD}\\
\hline
Increasing ratio & $67.4_{\pm0.22}$ & $70.4_{\pm0.21}$ & $67.9_{\pm0.70}$ & $63.5_{\pm1.04}$& $64.4_{\pm1.76}$\\
25\% masking &$67.3_{\pm0.29}$& $70.2_{\pm0.35}$ & $65.6_{\pm0.64}$  &$64.0_{\pm0.41}$& $61.4_{\pm2.16}$\\
50\% masking &$67.3_{\pm0.60}$& $70.2_{\pm0.43}$ & $66.2_{\pm1.07}$  &$63.6_{\pm0.68}$& $60.4_{\pm2.23}$\\
75\% masking  &$67.4_{\pm0.39}$& $70.4_{\pm0.39}$ & $65.3_{\pm1.15}$  &$63.6_{\pm0.84}$& $61.3_{\pm2.47}$\\
\hline
\end{tabular}
}
\caption{Linear probing accuracy (in \%) on MultiBench datasets after 100 epochs of self-supervised learning, comparing our adaptive masking approach with static masking ratios.}
\label{tab:table-4}
\end{table*}

We first evaluate the contribution of our asymmetric masking strategy using the synthetic Trifeature dataset. Table \ref{tab:table-3} compares performance under no masking, fixed masking ratios (25\%, 50\%, 75\%), and our progressively increasing masking schedule.  Without masking, COrAL achieves high accuracy on uniqueness and redundancy capture but fails to recover synergistic information, with performance only marginally above chance. 
This empirical evidence underscores that masking is essential for inducing an effective learning signal for synergy modeling. Introducing masking substantially improves synergy capture while preserving performance on redundant and unique components. 

On Trifeature, higher static masking ratios (e.g., 75\%) in the shared pathway slightly improve synergy capture, achieving $76.9_{\pm1.99}$\%, closely matching the performance of InfMasking ($77.0_{\pm4.22}$\%, Table \ref{tab:table-1}). Importantly, COrAL maintains strong unique information capture even under high masking levels. However, results on real-world MultiBench datasets (Table~\ref{tab:table-4}) indicate that high static masking ratios can degrade downstream performance and increase variance. We attribute this discrepancy to the artificially strong synergistic correlations in Trifeature, which may not reflect real-world modality interactions. Consequently, we adopt a progressively increasing masking schedule ($5\% \rightarrow 35\% \rightarrow 55\% \rightarrow 75\%$ over 25-epoch intervals) for COrAL, which provides stable performance across all datasets. This strategy balances synergy induction and representation stability, making it robust across heterogeneous multimodal settings.

\subsubsection{Loss weight sensitivity}
\begin{figure*}[t]
    \centering
    \includegraphics[width=\textwidth]{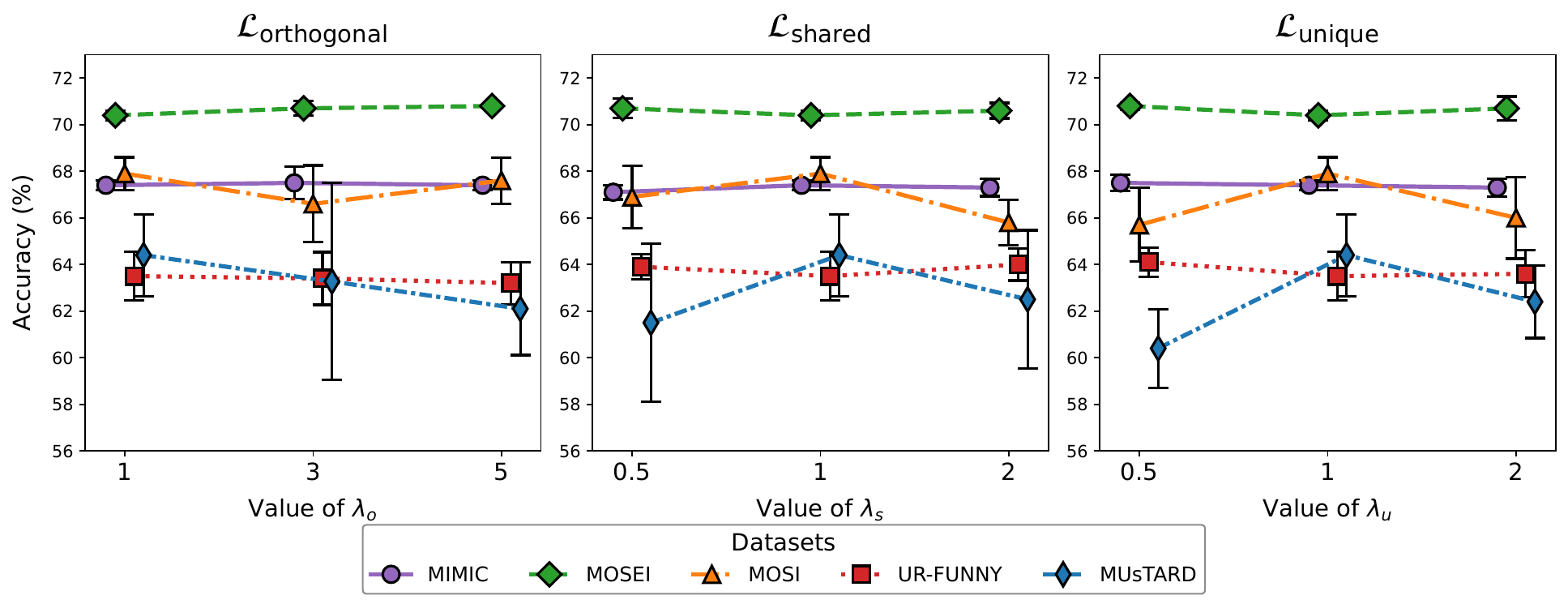}
    \caption{Sensitivity analysis of COrAL: Impact of loss weight variations on linear probing accuracy across MultiBench datasets after 100 epochs. For each of the three loss components $\mathcal{L}_{\text{orthogonal}}$, $\mathcal{L}_{\text{shared}}$, and $\mathcal{L}_{\text{unique}}$ we plot the changes in accuracy for different values of their $\lambda$, which weigh their influence on the overall loss $\mathcal{L}_{\text{COrAL}}$.}
    \label{fig:figure-4}
\end{figure*}

We next assess sensitivity to the loss weights $\lambda_s$, $\lambda_u$, and $\lambda_o$. Unless otherwise stated, we use $\lambda_s = \lambda_u = \lambda_o = 1$. To assess robustness, we independently halve or double $\lambda_s$ and $\lambda_u$ to examine the impact on each pathway, and increase $\lambda_o$ to 3 and 5, inspired by prior orthogonality-based formulations~\cite{Liu2023FOCAL}. In each experiment, we modify only one $\lambda$ weight at a time, keeping the other two at their default values. 
As shown in Figure \ref{fig:figure-4}, performance variations remain within 5\% across all datasets, indicating that COrAL is not overly sensitive to moderate reweighting. While small datasets exhibit slightly increased variance under altered weights, the default setting ($\lambda=1$ for all components) consistently yields low standard deviation and competitive accuracy. These findings confirm that COrAL does not require delicate hyperparameter tuning and that equal weighting provides a stable, general-purpose configuration.

\subsubsection{Unique pathway design}
\begin{table*}[t]
\centering
\resizebox{\textwidth}{!}{
\begin{tabular}{lcccccc}
\hline
\textbf{Unique Path Architecture} & \textbf{MIMIC} & \textbf{MOSEI} & \textbf{MOSI} & \textbf{UR-FUNNY} & \textbf{MUsTARD} & \textbf{Average} \\
\hline
MLP + pool (half size) (ours) & $67.4_{\pm0.22}$ &$70.4_{\pm0.21}$& ${67.9}_{\pm0.70}$ & ${63.5}_{\pm1.04}$& ${64.4}_{\pm1.76}$& ${66.7}_{\pm2.79}$ \\
MLP + pool (same size) & ${67.3}_{\pm0.52}$ & $70.3_{\pm0.43}$& ${66.1}_{\pm0.49}$ & ${63.6}_{\pm1.32}$& ${62.0}_{\pm1.34}$& ${65.9}_{\pm2.89}$ \\
Transformer (half size) & ${67.6}_{\pm0.46}$ &$70.4_{\pm0.24}$& ${66.6}_{\pm1.82}$ & ${63.9}_{\pm0.87}$& ${63.2}_{\pm1.69}$& ${{66.3}_{\pm2.61}}$ \\
\hline
\end{tabular}
}
\caption{Linear evaluation accuracy (in \%) on MultiBench datasets after 100 epochs of self-supervised learning, evaluating the impact of different architecture designs and embedding sizes for the unique pathway. We compare two architecture options for the unique pathway: a lightweight two-layer MLP with attention pooling (MLP + pool) and a Transformer-based approach, along with different embedding sizes relative to the shared pathway embedding.}
\label{tab:table-5}
\end{table*}
We further validate architectural choices for our unique pathway $F^{U}_i(\cdot)$ (Table \ref{tab:table-5}). We compare: (i) our proposed design with two-layer MLP with attention pooling, which outputs embeddings half the dimension of the shared pathway; (ii) a modified version of our architecture that maintains the same output dimensions as the shared embedding; and (iii) a lightweight Transformer (with 1-layer and 4 attention heads), with output dimensions reduced by half. 
As shown in Table \ref{tab:table-5}, our proposed design achieves comparable or superior performance across all datasets while maintaining lower complexity. Importantly, the Transformer-based variant does not yield consistent gains, confirming that increased architectural complexity does not necessarily translate into improved unique information extraction. This finding supports our lightweight design as an effective balance between representational capacity and efficiency.

\subsubsection{Disentangled embedding space}
\begin{figure*}[!t]
    \centering
    \includegraphics[width=0.7\textwidth]{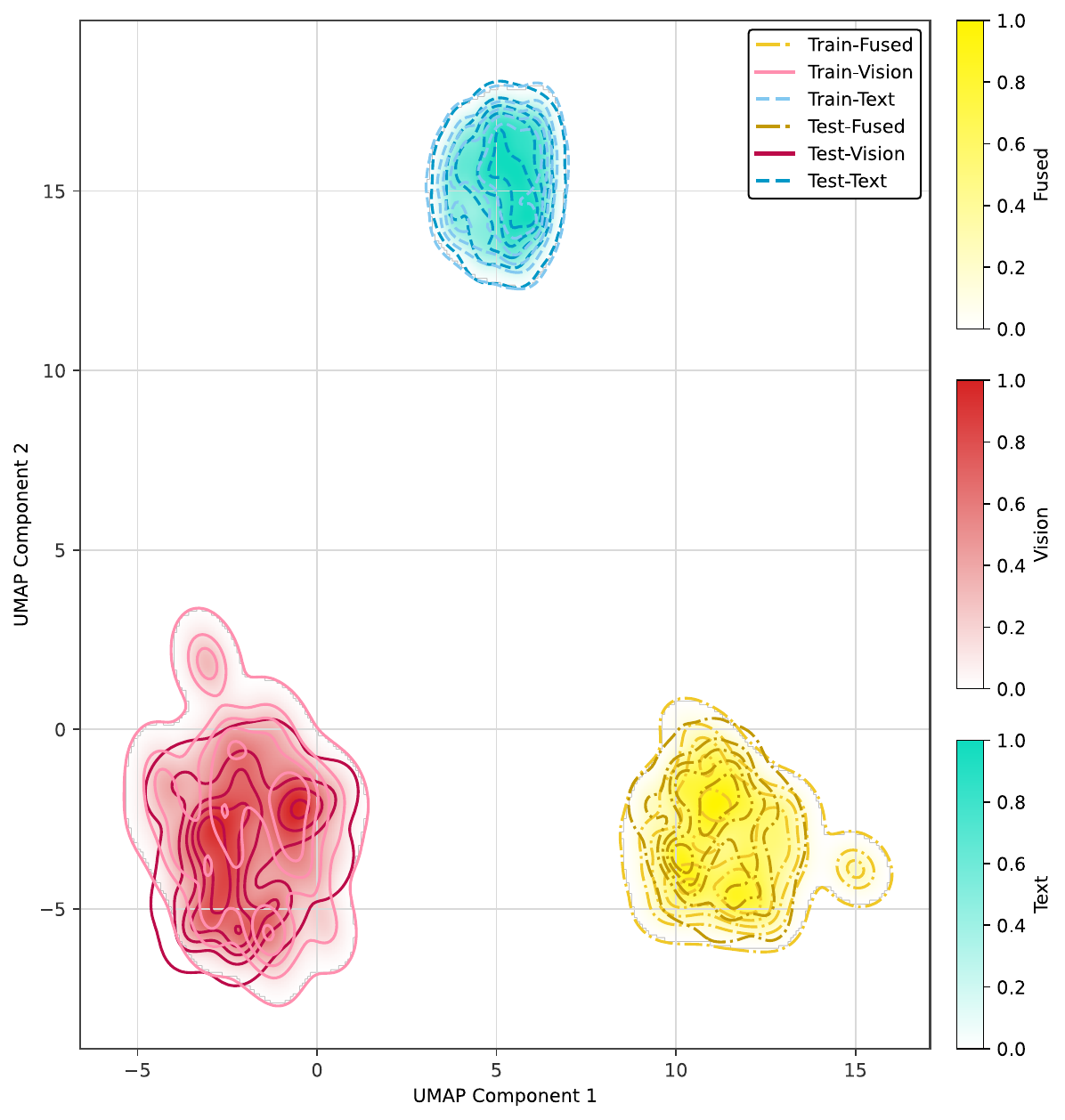}
    \caption{Visualization of the density of the shared and unique representations of visual and textual information from the MOSI dataset after projection into the embedding space using UMAP. Contour lines for training data density are shown in lighter colors, and contour lines for test data density are shown in darker colors.}
    \label{fig:figure-5}
\end{figure*}
COrAL explicitly enforces orthogonality between shared and unique representations, ensuring each captures distinct contents of information components. To verify whether this framework results in structural disentanglement, we project embeddings into a common subspace and visualize their distributions using UMAP (Figure \ref{fig:figure-5}). On MOSI (representative example), shared and unique embeddings form clearly separable subspaces, validating COrAL's ability to provide disentangled representations. Furthermore, test distributions align closely with training distributions, indicating that the learned structural separation generalizes beyond the training data. These results qualitatively validate the effectiveness of the orthogonality constraint and confirm that COrAL achieves meaningful disentanglement in the embedding space.

\subsubsection{Self-supervised finetuning}
While COrAL is optimized for self-supervised pretraining, downstream performance ultimately depends on fine-tuning. 
A key challenge in fine-tuning is preserving the decomposition of shared and unique information during supervised optimization, as naïve cross-entropy training can induce representation collapse. To mitigate this, we adopt a supervised contrastive fine-tuning strategy \cite{khosla2020supervised}. Positive pairs are constructed from same-class embeddings, and negative pairs from different labels, when computing $\mathcal{L}_{\text{shared}}$ and $\mathcal{L}_{\text{unique}}$. For datasets with high intra-class variability, we optionally restrict positive pairs to augmented views of the same sample, treating other same-class samples as neutral by exclusion from the loss. These relaxed constraints enable the model to preserve rich information that can capture dissimilar instances within a heterogeneous class label. 
Empirical tests indicated that freezing projection heads during fine-tuning and continuing to apply augmentations to generate two views for each sample, yields the best results.
During fine-tuning, the orthogonality loss $\mathcal{L}_{\text{orthogonal}}$ is continuously retained, although reducing its weight $\lambda_o$ may be beneficial depending on the downstream task. To further enhance synergistic information capture, we apply asymmetric masking at a lower ratio, as the model must now additionally perform cross-sample alignment of embeddings. Overall, fine-tuning introduces greater sensitivity to dataset characteristics, size, and task-specific objectives, making it difficult to prescribe universally optimal hyperparameters across all scenarios.

As a representative case, fine-tuning COrAL on CMU-MOSI after 100 epochs of pretraining (with 20\% masking, $\lambda_o=1$, and early stopping epochs: 46–89) achieves 75.3\% accuracy with 0.73\% standard deviation across five runs, yielding a 7.4\% absolute accuracy gain over the pretrained representation's linear evaluation. This result demonstrates that COrAL’s disentangled representations remain transferable and can be effectively adapted to downstream tasks without sacrificing structural integrity.
\section{Conclusion}
\label{sec:section_6}
This paper addresses the problem of structured information decomposition in multimodal representation learning. While many existing fusion frameworks primarily emphasize redundant information across modalities, they often overlook modality-unique and synergistic components or fail to separate them explicitly. To address this limitation, we proposed COrAL, a multimodal framework that jointly models redundant, modality-specific, and synergistic information within a unified and principled architecture. 
COrAL combines a dual-path design with orthogonality constraints to enforce separation between shared and unique subspaces, thereby reducing cross-component interference. In addition, an asymmetric masking mechanism is integrated into the shared pathway to promote synergistic feature learning without introducing auxiliary losses or additional representational branches. This design enables explicit and controlled decomposition of multimodal information while maintaining architectural efficiency.

Experimental results on both synthetic data and MultiBench benchmarks demonstrate that COrAL effectively captures the three targeted information components and produces stable, competitive downstream performance. On synthetic experiments, the framework shows improved modeling of modality-unique information compared to existing methods. On real-world datasets, COrAL achieves state-of-the-art or comparable accuracy with consistently low variance, indicating robust and reliable representation learning.

Future work will focus on extending the framework to settings involving more than two modalities and on developing scalable formulations of the orthogonality and masking mechanisms for higher-order interactions. Overall, this work contributes a structured and empirically validated approach to multimodal fusion, advancing the principled modeling of redundant, unique, and synergistic information in complex multimodal systems.

\section*{CRediT authorship contribution statement}
\textbf{Carolin Cissée:} Conceptualization, Methodology, Software, Validation, Investigation, Writing - Original Draft, Visualization; 
\textbf{Raneen Younis:} Writing - Review \& Editing, Project administration;
\textbf{Zahra Ahmadi:} Conceptualization, Validation, Writing - Review \& Editing, Supervision, Funding acquisition.

\section*{Data availability}
The Multibench datasets \cite{Liang2021MultiBench} CMU-MOSEI \cite{Zadeh2018MOSEI}, CMU-MOSI \cite{Zadeh2016MOSI}, UR-FUNNY \cite{Hasan2019URFUNNY}, and MUsTARD \cite{Castro2019MUsTARD} are publicly available, except for the MIMIC III dataset \cite{Johnson2016MIMIC3, Liang2021MultiBench}, which is available upon providing proof of the required credentials. The code for generating the synthetic data is made available by Dufumier et al. \cite{Dufumier2024CoMM}.

\section*{Declaration of competing interest}
The authors declare that they have no known competing financial interests or personal relationships that could have appeared to influence the work reported in this article.

\section*{Acknowledgment}
This work was supported by the Federal Ministry of Research, Technology and Space of Germany [grant number 16IS24059]. We acknowledge the Hannover Medical School for providing MHH-HPC resources that have contributed to the results reported in this paper. 


\bibliographystyle{elsarticle-num} 
\bibliography{paper-bib}

\end{document}